\documentclass[sigconf]{acmart}
\usepackage{subfigure}
\usepackage{relsize}
\usepackage{natbib}
\renewcommand\footnotetextcopyrightpermission[1]{} 

\setcopyright{rightsretained}

\acmDOI{}

\acmISBN{}

\acmConference{SysML Conference}
\acmYear{2018}
\copyrightyear{2018}

\acmArticle{4}
\acmPrice{15.00}

\renewcommand{\baselinestretch}{0.97}
\begin{document}
\title{Not All Ops Are Created Equal!}


\author{Liangzhen Lai}
\affiliation{
  \institution{Arm Inc.}
}
\email{liangzhen.lai@arm.com}

\author{Naveen Suda}
\affiliation{
  \institution{Arm Inc.}
}
\email{naveen.suda@arm.com}

\author{Vikas Chandra}
\affiliation{
  \institution{Arm Inc.}
}
\email{vikas.chandra@arm.com}
\email{}

\begin{abstract}
Efficient and compact neural network models are
essential for enabling the deployment on mobile and embedded devices.
In this work, we point out that typical design metrics 
for gauging the efficiency of neural network
architectures -- total number of operations and
parameters -- are not sufficient. These metrics may not accurately
correlate with the actual deployment metrics such as energy and  memory footprint.
We show that throughput and energy varies by up to 5X across 
different neural network operation types on an off-the-shelf Arm Cortex-M7 microcontroller. 
Furthermore, we show that the memory required for activation data also need to be
considered, apart from the model parameters, for network architecture 
exploration studies. 
\end{abstract}

\maketitle

\section{Introduction}

Exploring efficient neural network (NN) architectures targeted for mobile and embedded
devices with constrained energy and memory resources has been the recent trend 
in machine learning research~\cite{mobilenet,shufflenet,squeezenet,rastegari2016xnor,zhang2017hello}.
Most research use the number of operations (Ops) and/or parameters (i.e., weights) 
as the metrics for evaluating the model complexity and compactness. 
While these metrics are sufficient when comparing significantly different 
NN models (e.g. AlexNet~\cite{krizhevsky2012imagenet} vs. 
MobileNets~\cite{mobilenet}), they may not be accurate enough for comparing networks 
whose complexity and sizes are similar.
Furthermore, as research shifts towards fine-grained optimization,
e.g., network architecture search~\cite{zoph2016neural,baker2016designing} and
hyperparameter search~\cite{zhang2017hello,stamoulis2017hyperpower},
reductions in Ops or parameters may not always improve the network efficiency.


Energy per inference and total memory footprint are two main system
metrics to be considered for deploying NN based solutions
on resource constrained devices. In this work, we show examples that NNs with similar
network design metrics can have very different deployment metrics when running
on resource constrained devices like microcontrollers.
In particular, we show:
\begin{itemize}
\item Throughput and energy efficiency for different types of NN operations can vary
by up to 5X. This can result in 30\% difference in runtime and energy for NNs with
similar Ops and accuracy.
\item 
Different operations with same amount of weights can have varying amount of 
activation data, and thus different memory footprint. 
This may not be an issue for large-scale systems, but is critical for devices
with limited memory.
\end{itemize}

All experiments are performed using optimized neural network kernels
in CMSIS-NN~\cite{lai2018cmsis}.
The delay/power results are measured on a NUCLEO-F746ZG mbed
development board~\cite{nucleo_m7}, which has an Arm
Cortex-M7 core (running at 216 MHz), 1 MB flash and 320 KB SRAM.

\section{Energy per Inference} 

\begin{figure}[t]
\includegraphics[clip, trim=2cm 9.5cm 2cm 9.5cm, width = 0.9\columnwidth]{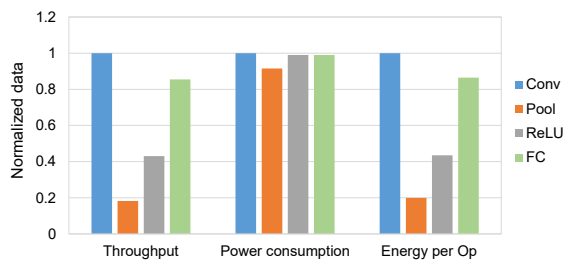}
\caption{Normalized throughput and energy of different type of NN
operations in a CNN for CIFAR-10 dataset.}
\label{fig:ops}
\end{figure}

Energy consumption per inference is a crucial metric that 
determines the battery life of an embedded system and it is imperative 
that NN models are optimized for energy efficiency. 
Typically, number of Ops is considered as a proxy for the energy
consumption per inference, but the type of operations also has huge impact on 
the energy. For example, 
Fig.~\ref{fig:ops} shows the normalized throughput, power consumption and 
energy per Op of different NN operation types of the convolutional
neural network (CNN) for CIFAR-10 dataset from Caffe examples~\cite{jia2014caffe}. 
The results show that throughput (i.e. Ops/s) can vary by 5X across different operation 
types, but average power consumption remains almost same.
This implies that the
overall energy consumption depends mostly on the throughput.
Among all the operation types, max pooling is particularly slow because it is based
on comparisons (i.e. branch) rather than computations.
However, in a typical NN, convolution and fully-connected (FC) layers constitute 
more than 90\% of the operations. These layers achieve good throughput by effectively utilizing
the SIMD Multiply-Accumulate (MAC) instructions. 

\begin{figure}[b]
\includegraphics[clip, trim=2cm 9.5cm 2cm 9.5cm, width = 0.9\columnwidth]{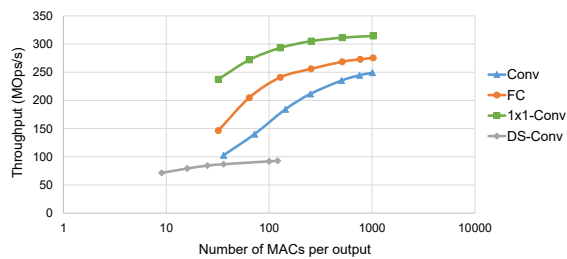}
\caption{Throughput variation with number of MACs per output for different types of
NN operations. Depthwise separable convolution (DS-Conv) typically has 
much less MACs per output compared to other operation types.}
\label{fig:throughput_trends}
\end{figure}

Fig.~\ref{fig:throughput_trends} shows the throughput of different MAC based
NN operations. Since the throughput depends heavily on the layer dimensions,
we use the number of MAC operations per output to represent the effectiveness 
of SIMD MAC instructions. In this case, the difference between operation types 
represents the relative overhead of fetching the MAC operands.
In general, convolution is slower than fully-connected layer because of additional
\textit{im2col} overhead. However, 1x1 convolution does not require \textit{im2col}.
It uses matrix-matrix multiplication (GEMM) style of computations, which is faster than 
matrix-vector multiplication (GEMV) style of computations used in fully-connected layer 
due to better data reuse.
Among all operation types, depthwise separable convolution (DS-Conv) is the slowest as it has
higher \textit{im2col} overhead and typically lower MACs per output.

\begin{figure}[t]
\includegraphics[clip, trim=1cm 5.5cm 1cm 5.5cm, width = 0.9\columnwidth]{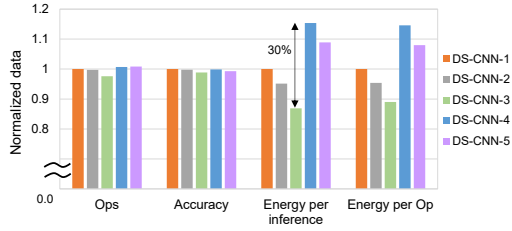}
\caption{Normalized energy consumption of 5 DS-CNN models with similar accuracy and number of Ops.}
\label{fig:network_ops}
\end{figure}

Understanding the throughput differences between operation types 
is crucial for designing efficient NN architectures.
Fig.~\ref{fig:network_ops} shows the normalized energy consumption, number
of Ops and accuracy of 5 DS-CNN 
models~\cite{zhang2017hello} with different number of layers and 
features per layer, trained on Google speech commands dataset
~\cite{google_dataset}.  
It shows that the energy per inference varies by as much as 30\% across these 
models although they come from the same NN architecture family and
have similar accuracy and total number of Ops.

\begin{figure}[h]
\subfigure{\includegraphics[clip, trim=2.4cm 4cm 2cm 4cm, width=0.49\columnwidth]{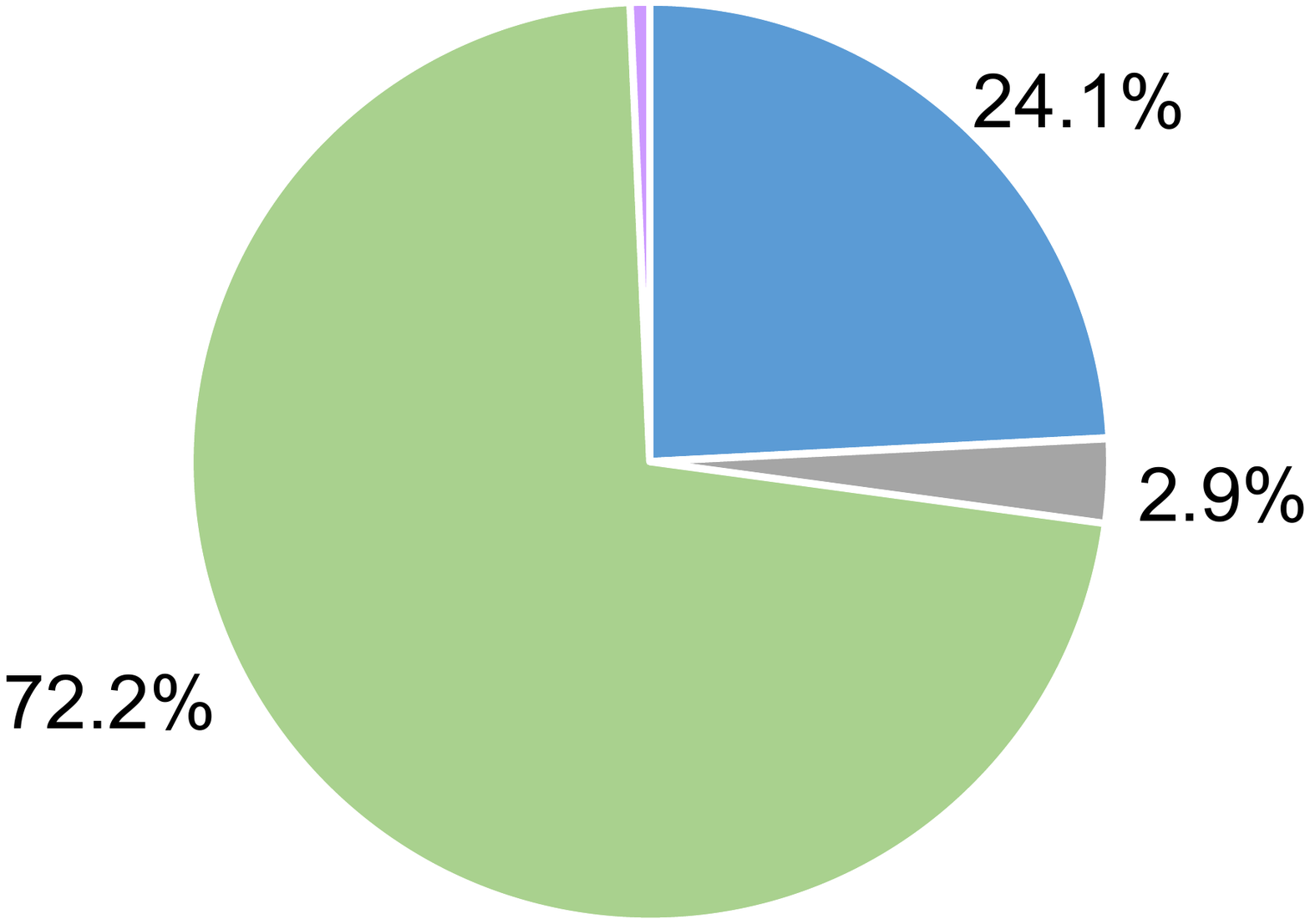}}
\subfigure{\includegraphics[clip, trim=2.4cm 4cm 2cm 4cm, width=0.49\columnwidth]{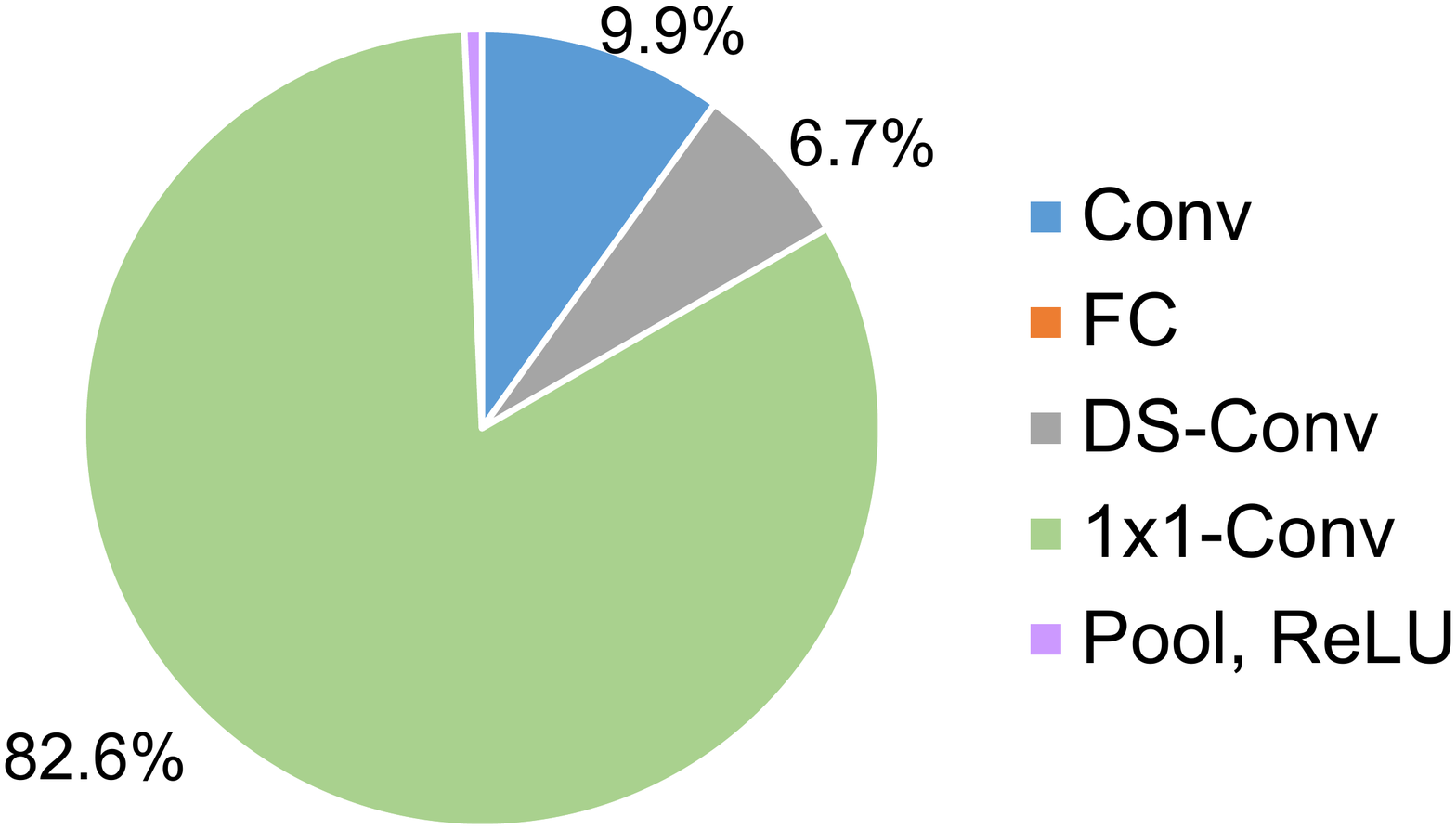}}
\caption{Distribution of different operation types in DS-CNN-3 (left) and DS-CNN-4 (right)
models shown in Fig.~\ref{fig:network_ops}.}
\label{fig:pie}
\end{figure}

The distributions of the different operation types of DS-CNN-3 and DS-CNN-4 models 
are shown in Fig.~\ref{fig:pie}. 
Compared to the DS-CNN-3 model, DS-CNN-4 has higher proportion of DS-Conv Ops, 
which has substantially lower throughput compared to other operation types
as shown in Fig.~\ref{fig:throughput_trends}.
This results in 30\% reduction in the overall throughput and hence energy efficiency.
Using Ops as a metric without considering the throughput of different operation types on the 
actual hardware may lead to sub-optimal efficiency. 

When performing fine-grained NN optimization, the operation type
and dimensions should be considered for evaluating the network efficiency.
The results we show in this work are based on a general purpose processor and
the operation characteristics for other platforms
(e.g. GPU, FPGA, DSP, accelerator) can be very different.
Performance for different operation types, similar to results in
Fig.~\ref{fig:throughput_trends}, can be pre-characterized for the 
target hardware platform and used for estimating the network efficiency.

\section{Memory Footprint}
System memory size is the other important limiting factor for
running NNs on resource constrained devices.
For example, typical microcontroller SoC have 100 KB - 1 MB of flash
(to store program binary and model weights) and 10-300 KB of SRAM
(to store the activation data).

\begin{figure}[h]
\includegraphics[clip, trim=2cm 10cm 2cm 10cm, width = 0.9\columnwidth]{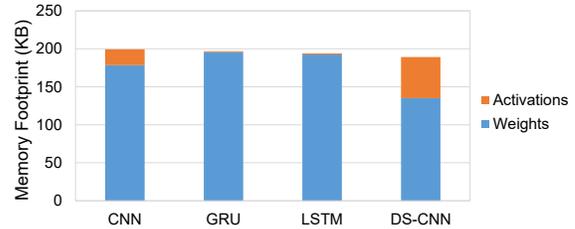}
\caption{Memory footprint (total weights and maximum concurrent activation data) 
breakdown for four different types of models from~\cite{zhang2017hello}.
}
\label{fig:footprint}
\end{figure}

The number of model parameters, which can be used as metric to quantify the 
compactness of a NN model, determines whether the model fits in the flash or not.
However, it may not be a good metric for representing the
total memory footprint, as it does not consider the activation data typically stored
in the SRAM.
The amount of activation
data can be a significant part of the total memory footprint and will
depend on the operation type as well. 
For example, Fig.~\ref{fig:footprint} shows the memory footprint of four NN models 
for the keyword spotting application from~\cite{zhang2017hello}. 
The size of maximum concurrent 
activation data varies between 1\% to 30\% of the total memory footprint.

Apart from operation types, NN topology can also affect
the size of maximum concurrent activation data. Regular feed-forward
network need to store only the input and output data of the current layer. 
If there are other feed-forward connections,
such as in DenseNet~\cite{huang2016densely}, the total number of concurrent activation data will
increase. Also, some networks generated by automatic network architecture search
can have many feed-forward connections~\cite{zoph2016neural}, which 
can substantially ($\sim$10X) increase the size of the activation data.

The network optimization target in NN architecture exploration should be the
total memory footprint instead of the number of parameters.
During NN architecture exploration, total memory
footprint can be estimated by the sum of the size of maximum concurrent activation 
data and the size of weight parameters.
The maximum concurrent activation data can be obtained from the
network graph and the execution order.
The concurrent activation data, when executing an operation, will include
the operation input, output, as well as other activation data (i.e. feed-forward edges)
that are needed for the operation.

\section{Conclusion}
In this work, we show that the NN operation type
has significant impact on system efficiency.
The commonly used network design metrics -- 
number of operations and parameters -- need to be rethought as
they may not accurately correlate with system design metrics such as
energy efficiency and memory footprint. 
Experimental results on an off-the-shelf Arm Cortex-M microcontroller show 
that the energy per operation can vary up to 5X for different NN operation types.
Network activation data, which are typically 
overlooked can also contribute to up to 30\% of the total memory footprint. 
The network architecture exploration should account for both energy
efficiency as well as total memory footprint to make the inference
more efficient on resource constrained devices.

\renewcommand{\baselinestretch}{1.00}


\begin{thebibliography}{14}


\ifx \showCODEN    \undefined \def \showCODEN     #1{\unskip}     \fi
\ifx \showDOI      \undefined \def \showDOI       #1{#1}\fi
\ifx \showISBNx    \undefined \def \showISBNx     #1{\unskip}     \fi
\ifx \showISBNxiii \undefined \def \showISBNxiii  #1{\unskip}     \fi
\ifx \showISSN     \undefined \def \showISSN      #1{\unskip}     \fi
\ifx \showLCCN     \undefined \def \showLCCN      #1{\unskip}     \fi
\ifx \shownote     \undefined \def \shownote      #1{#1}          \fi
\ifx \showarticletitle \undefined \def \showarticletitle #1{#1}   \fi
\ifx \showURL      \undefined \def \showURL       {\relax}        \fi
\providecommand\bibfield[2]{#2}
\providecommand\bibinfo[2]{#2}
\providecommand\natexlab[1]{#1}
\providecommand\showeprint[2][]{arXiv:#2}

\bibitem[\protect\citeauthoryear{??}{nuc}{}]%
        {nucleo_m7}
 \bibinfo{year}{}\natexlab{}
\newblock \bibinfo{title}{NUCLEO-F746ZG development board}.
\newblock
  \bibinfo{howpublished}{http://www.st.com/en/evaluation-tools/nucleo-f746zg.html}.
\newblock


\bibitem[\protect\citeauthoryear{Baker, Gupta, Naik, and Raskar}{Baker
  et~al\mbox{.}}{2016}]%
        {baker2016designing}
\bibfield{author}{\bibinfo{person}{Bowen Baker}, \bibinfo{person}{Otkrist
  Gupta}, \bibinfo{person}{Nikhil Naik}, {and} \bibinfo{person}{Ramesh
  Raskar}.} \bibinfo{year}{2016}\natexlab{}.
\newblock \showarticletitle{Designing neural network architectures using
  reinforcement learning}.
\newblock \bibinfo{journal}{\emph{arXiv preprint arXiv:1611.02167}}
  (\bibinfo{year}{2016}).
\newblock


\bibitem[\protect\citeauthoryear{Howard, Zhu, Chen, Kalenichenko, Wang, Weyand,
  Andreetto, and Adam}{Howard et~al\mbox{.}}{2017}]%
        {mobilenet}
\bibfield{author}{\bibinfo{person}{Andrew~G Howard}, \bibinfo{person}{Menglong
  Zhu}, \bibinfo{person}{Bo Chen}, \bibinfo{person}{Dmitry Kalenichenko},
  \bibinfo{person}{Weijun Wang}, \bibinfo{person}{Tobias Weyand},
  \bibinfo{person}{Marco Andreetto}, {and} \bibinfo{person}{Hartwig Adam}.}
  \bibinfo{year}{2017}\natexlab{}.
\newblock \showarticletitle{Mobilenets: Efficient convolutional neural networks
  for mobile vision applications}.
\newblock \bibinfo{journal}{\emph{arXiv preprint arXiv:1704.04861}}
  (\bibinfo{year}{2017}).
\newblock


\bibitem[\protect\citeauthoryear{Huang, Liu, Weinberger, and van~der
  Maaten}{Huang et~al\mbox{.}}{2016}]%
        {huang2016densely}
\bibfield{author}{\bibinfo{person}{Gao Huang}, \bibinfo{person}{Zhuang Liu},
  \bibinfo{person}{Kilian~Q Weinberger}, {and} \bibinfo{person}{Laurens van~der
  Maaten}.} \bibinfo{year}{2016}\natexlab{}.
\newblock \showarticletitle{Densely connected convolutional networks}.
\newblock \bibinfo{journal}{\emph{arXiv preprint arXiv:1608.06993}}
  (\bibinfo{year}{2016}).
\newblock


\bibitem[\protect\citeauthoryear{Iandola, Han, Moskewicz, Ashraf, Dally, and
  Keutzer}{Iandola et~al\mbox{.}}{2016}]%
        {squeezenet}
\bibfield{author}{\bibinfo{person}{Forrest~N Iandola}, \bibinfo{person}{Song
  Han}, \bibinfo{person}{Matthew~W Moskewicz}, \bibinfo{person}{Khalid Ashraf},
  \bibinfo{person}{William~J Dally}, {and} \bibinfo{person}{Kurt Keutzer}.}
  \bibinfo{year}{2016}\natexlab{}.
\newblock \showarticletitle{SqueezeNet: AlexNet-level accuracy with 50x fewer
  parameters and< 0.5 MB model size}.
\newblock \bibinfo{journal}{\emph{arXiv preprint arXiv:1602.07360}}
  (\bibinfo{year}{2016}).
\newblock


\bibitem[\protect\citeauthoryear{Jia, Shelhamer, Donahue, Karayev, Long,
  Girshick, Guadarrama, and Darrell}{Jia et~al\mbox{.}}{2014}]%
        {jia2014caffe}
\bibfield{author}{\bibinfo{person}{Yangqing Jia}, \bibinfo{person}{Evan
  Shelhamer}, \bibinfo{person}{Jeff Donahue}, \bibinfo{person}{Sergey Karayev},
  \bibinfo{person}{Jonathan Long}, \bibinfo{person}{Ross Girshick},
  \bibinfo{person}{Sergio Guadarrama}, {and} \bibinfo{person}{Trevor Darrell}.}
  \bibinfo{year}{2014}\natexlab{}.
\newblock \showarticletitle{Caffe: Convolutional architecture for fast feature
  embedding}. In \bibinfo{booktitle}{\emph{Proceedings of the 22nd ACM
  international conference on Multimedia}}. ACM, \bibinfo{pages}{675--678}.
\newblock


\bibitem[\protect\citeauthoryear{Krizhevsky, Sutskever, and Hinton}{Krizhevsky
  et~al\mbox{.}}{2012}]%
        {krizhevsky2012imagenet}
\bibfield{author}{\bibinfo{person}{Alex Krizhevsky}, \bibinfo{person}{Ilya
  Sutskever}, {and} \bibinfo{person}{Geoffrey~E Hinton}.}
  \bibinfo{year}{2012}\natexlab{}.
\newblock \showarticletitle{Imagenet classification with deep convolutional
  neural networks}. In \bibinfo{booktitle}{\emph{Advances in neural information
  processing systems}}. \bibinfo{pages}{1097--1105}.
\newblock


\bibitem[\protect\citeauthoryear{Lai, Suda, and Chandra}{Lai
  et~al\mbox{.}}{2018}]%
        {lai2018cmsis}
\bibfield{author}{\bibinfo{person}{Liangzhen Lai}, \bibinfo{person}{Naveen
  Suda}, {and} \bibinfo{person}{Vikas Chandra}.}
  \bibinfo{year}{2018}\natexlab{}.
\newblock \showarticletitle{CMSIS-NN: Efficient Neural Network Kernels for Arm
  Cortex-M CPUs}.
\newblock \bibinfo{journal}{\emph{arXiv preprint arXiv:1801.06601}}
  (\bibinfo{year}{2018}).
\newblock


\bibitem[\protect\citeauthoryear{Rastegari, Ordonez, Redmon, and
  Farhadi}{Rastegari et~al\mbox{.}}{2016}]%
        {rastegari2016xnor}
\bibfield{author}{\bibinfo{person}{Mohammad Rastegari},
  \bibinfo{person}{Vicente Ordonez}, \bibinfo{person}{Joseph Redmon}, {and}
  \bibinfo{person}{Ali Farhadi}.} \bibinfo{year}{2016}\natexlab{}.
\newblock \showarticletitle{Xnor-net: Imagenet classification using binary
  convolutional neural networks}. In \bibinfo{booktitle}{\emph{European
  Conference on Computer Vision}}. Springer, \bibinfo{pages}{525--542}.
\newblock


\bibitem[\protect\citeauthoryear{Stamoulis, Cai, Juan, and
  Marculescu}{Stamoulis et~al\mbox{.}}{2017}]%
        {stamoulis2017hyperpower}
\bibfield{author}{\bibinfo{person}{Dimitrios Stamoulis}, \bibinfo{person}{Ermao
  Cai}, \bibinfo{person}{Da-Cheng Juan}, {and} \bibinfo{person}{Diana
  Marculescu}.} \bibinfo{year}{2017}\natexlab{}.
\newblock \showarticletitle{HyperPower: Power-and Memory-Constrained
  Hyper-Parameter Optimization for Neural Networks}.
\newblock \bibinfo{journal}{\emph{arXiv preprint arXiv:1712.02446}}
  (\bibinfo{year}{2017}).
\newblock


\bibitem[\protect\citeauthoryear{Warden}{Warden}{2017}]%
        {google_dataset}
\bibfield{author}{\bibinfo{person}{Pete Warden}.}
  \bibinfo{year}{2017}\natexlab{}.
\newblock \showarticletitle{Speech Commands: A public dataset for single-word
  speech recognition.}
\newblock \bibinfo{journal}{\emph{Dataset available from
  http://download.tensorflow.org/data/speech\_commands\_v0.01.tar.gz}}
  (\bibinfo{year}{2017}).
\newblock


\bibitem[\protect\citeauthoryear{Zhang, Zhou, Lin, and Sun}{Zhang
  et~al\mbox{.}}{2017b}]%
        {shufflenet}
\bibfield{author}{\bibinfo{person}{Xiangyu Zhang}, \bibinfo{person}{Xinyu
  Zhou}, \bibinfo{person}{Mengxiao Lin}, {and} \bibinfo{person}{Jian Sun}.}
  \bibinfo{year}{2017}\natexlab{b}.
\newblock \showarticletitle{Shufflenet: An extremely efficient convolutional
  neural network for mobile devices}.
\newblock \bibinfo{journal}{\emph{arXiv preprint arXiv:1707.01083}}
  (\bibinfo{year}{2017}).
\newblock


\bibitem[\protect\citeauthoryear{Zhang, Suda, Lai, and Chandra}{Zhang
  et~al\mbox{.}}{2017a}]%
        {zhang2017hello}
\bibfield{author}{\bibinfo{person}{Yundong Zhang}, \bibinfo{person}{Naveen
  Suda}, \bibinfo{person}{Liangzhen Lai}, {and} \bibinfo{person}{Vikas
  Chandra}.} \bibinfo{year}{2017}\natexlab{a}.
\newblock \showarticletitle{Hello Edge: Keyword Spotting on Microcontrollers}.
\newblock \bibinfo{journal}{\emph{arXiv preprint arXiv:1711.07128}}
  (\bibinfo{year}{2017}).
\newblock


\bibitem[\protect\citeauthoryear{Zoph and Le}{Zoph and Le}{2016}]%
        {zoph2016neural}
\bibfield{author}{\bibinfo{person}{Barret Zoph} {and} \bibinfo{person}{Quoc~V
  Le}.} \bibinfo{year}{2016}\natexlab{}.
\newblock \showarticletitle{Neural architecture search with reinforcement
  learning}.
\newblock \bibinfo{journal}{\emph{arXiv preprint arXiv:1611.01578}}
  (\bibinfo{year}{2016}).
\newblock


\end{thebibliography}


\end{document}